\documentclass[runningheads]{llncs}
\usepackage{graphicx}
\usepackage{comment}
\usepackage{amsmath,amssymb} 
\usepackage{color}

\usepackage{algorithm}
\usepackage{algorithmicx}
\usepackage{algpseudocode}
\usepackage{amsmath}
\usepackage[width=122mm,left=12mm,paperwidth=146mm,height=193mm,top=12mm,paperheight=217mm]{geometry}
\usepackage{cases}
\usepackage{subfigure}
\usepackage{slashed}

\begin{document}

\pagestyle{headings}
\mainmatter
\def\ECCVSubNumber{6021}  

\title{Boundary Content Graph Neural Network for Temporal Action Proposal Generation} 

\titlerunning{BC-GNN}
%
\author{Yueran Bai\inst{1}$^{* \dagger}$ \and 
Yingying Wang\inst{2}$^{*}$ \and 
Yunhai Tong\inst{1}$^{\ddag}$ \and 
Yang Yang \inst{2} \and 
Qiyue Liu \inst{2} \and 
Junhui Liu \inst{2}$^{\ddag}$ 
}

\authorrunning{Yueran bai, Yingying Wang, and et al.}
%
\institute{Key Laboratory of Machine Perception (MOE), School of EECS, Peking University \\ \and
iQIYI, Inc. \\
\email{\{baiyueran,yhtong\}@pku.edu.cn \{wangyingying02,andyang,liuqiyue,liujunhui\}@qiyi.com}}

\maketitle

\begin{abstract}
Temporal action proposal generation plays an important role in video action understanding, which requires localizing high-quality action content precisely. However, generating temporal proposals with both precise boundaries and high-quality action content is extremely challenging. To address this issue, we propose a novel Boundary Content Graph Neural Network (BC-GNN) to model the insightful relations between the boundary and action content of temporal proposals by the graph neural networks. In BC-GNN, the boundaries and content of temporal proposals are taken as the nodes and edges of the graph neural network, respectively, where they are spontaneously linked. Then a novel graph computation operation is proposed to update features of edges and nodes. After that, one updated edge and two nodes it connects are used to predict boundary probabilities and content confidence score, which will be combined to generate a final high-quality proposal. Experiments are conducted on two mainstream datasets: ActivityNet-1.3 and THUMOS14. Without the bells and whistles, BC-GNN outperforms previous state-of-the-art
methods in both temporal action proposal and temporal action detection tasks.
\end{abstract}
\renewcommand{\thefootnote}{}
\footnotetext{$*$ Equal contributions}
\footnotetext{$\ddag$ Corresponding author}
\footnotetext{$\dagger$ Work was done during an internship in iQIYI, Inc.}
\section{Introduction}
Temporal action proposal generation becomes an active research topic in recent years, as it is a fundamental step for untrimmed video understanding tasks, such as temporal action detection and video analysis. A useful action proposal method could distinguish the activities we are interested in, so that only intervals containing visual information indicating activity categories can be retrieved. Although extensive studies have been carried out in the past, generating temporal proposals with both precise boundaries and rich action content remains a challenge\cite{wang2014action,oneata2014lear,SST,sw2,sw3,SSAD,BSN,BMN}.

Some existing methods \cite{wang2014action,oneata2014lear,SST,sw2,sw3,SSAD} are proposed to generate candidate proposals by sliding multi-scale temporal windows in videos with regular interval or designing multiple temporal anchor instances for temporal feature maps. Since the lengths of windows and anchors are fixed and set previously, these methods cannot generate proposals with precise boundaries and lack flexibility to retrieve action instances of varies temporal durations.

Recent works \cite{BSN,BMN} aim to generate higher quality proposals. \cite{BSN} adopts a ``local to global'' fashion to retrieve proposals. In the first, temporal boundaries are achieved by evaluating boundary confidence of every location of the video feature sequence. Then, content feature between boundaries of each proposal is used to generate content confidence score of proposal. \cite{BMN} proposes an end-to-end pipeline, in which confidence score of boundaries and content of densely distributed proposals are generated simultaneously. Although these works can generate proposals with higher quality, they ignore to make explicit use of interaction between boundaries and content.


\begin{figure}
\centering
\includegraphics[width=.72 \linewidth]{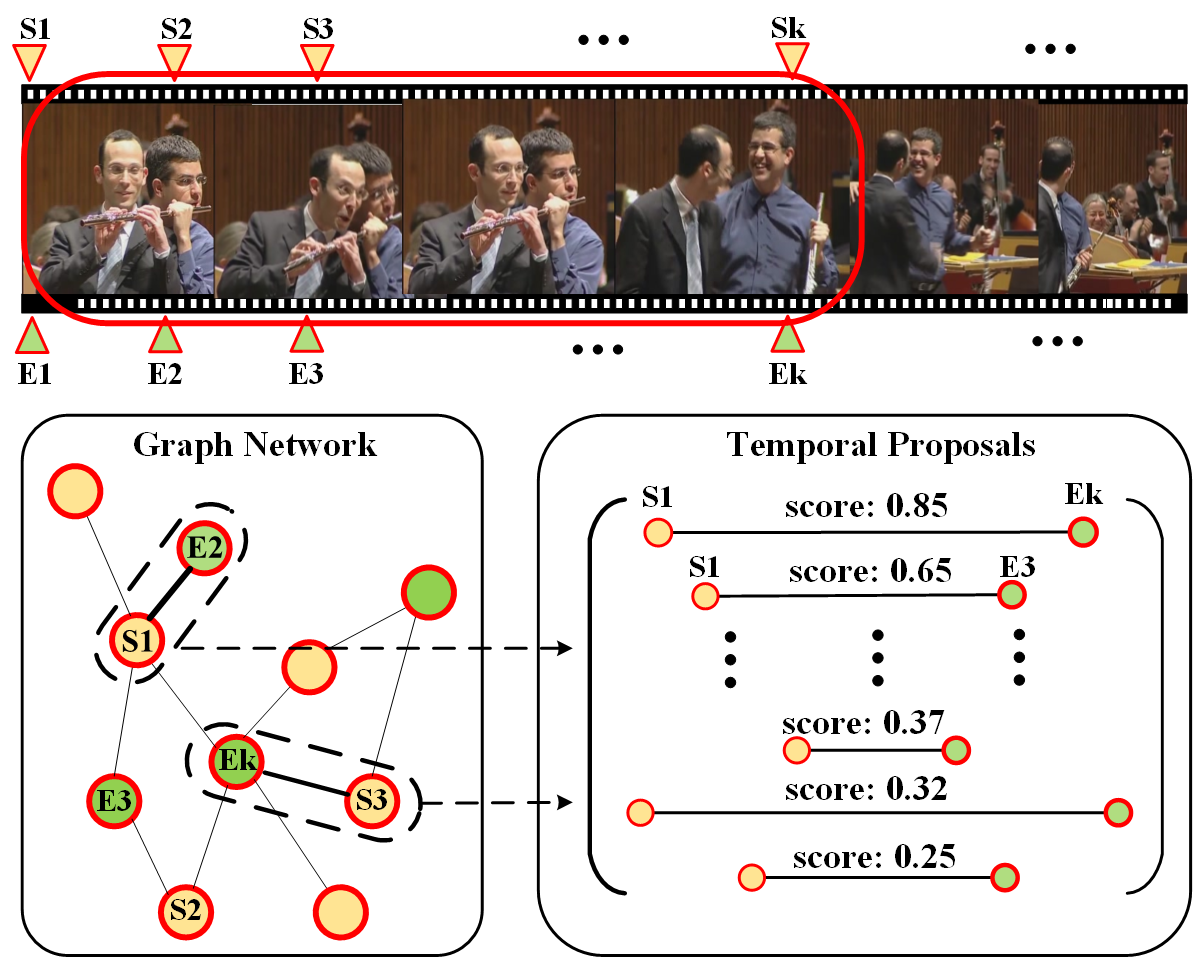}
\setlength{\belowcaptionskip}{-.0cm}
\caption{ Schematic depiction of the proposed approach. The red box denotes an action instance in a video. We regard temporal locations with regular interval as start locations and end locations for video segments. Start locations $S$ and end locations $E$ are regarded as nodes. Only when the location of $S$ is before $E$, we define the content between them as an edge to connect them. Then, a novel graph reasoning operation is applied to enable the relationship between nodes and edges. Finally, two nodes and the edge connected them form a temporal proposal.}
\label{introduct}
\end{figure}
To address this drawback, we propose Boundary Content Graph Neural Network (BC-GNN), which uses a graph neural network to model interaction between boundaries and content of proposals. As shown in Fig \ref{introduct}, a graph neural network links boundaries and content into a whole. For the graph of each video, the nodes denote temporal locations, while the edges between nodes are defined based on content between these locations. This graph enables information exchanging between nodes and edges to generate more dependable boundary probabilities and content confidence scores. In our proposed framework, a graph neural network is constructed to link boundaries and content of temporal proposals firstly.  Then a novel graph computation operation is proposed to update features of edges and nodes. After that, one updated edge and two nodes it connects are used to product boundary probabilities and content confidence score, which are combined to generate a candidate proposal.

In summary, the main contributions of our work are three folds:

(1) We propose a new approach named Boundary Content Graph Neural Network (BC-GNN) based on the graph neural network to enable the relationship between boundary probability predictions and confidence evaluation procedures.

(2) We introduce a novel graph reasoning operation in BC-GNN to update attributes of the edges and nodes in the boundary content graph.

(3) Experiments in different datasets demonstrate that our method outperforms other existing state-of-the-art methods in both temporal action proposal generation task and temporal action detection task.

\section{Related work}

\noindent\textbf{Action Recognition.}
Recognizing action classes in trimmed videos is a both basic and significant task for the purpose of video understanding. Traditional approaches are mostly based on hand-crafted feature \cite{one,two,three,four}. As the progress of Convolutional Neural Networks (CNN) in recent years, CNN based methods are widely adopted in action recognition and achieve superior performance. One type of these methods \cite{five,six} focus on combining multiple data modalities. Furthermore, other methods attempt to exploit the spatial-temporal feature by using 3D convolution operation \cite{eight,nine,ten}. The feature sequence extracted by action recognition models can be used as the input feature sequence of our network framework to analyze long and untrimmed video.

\noindent\textbf{Graph Neural Network.} Graph Neural Networks(GNNs) are proposed to handle graph-structured data with deep learning. With the development of deep learning, different kinds of GNNs appear one after another. \cite{Kipf_GCN} proposes the Graph Convolutional Networks(GCNs), which defines convolutions on the non-grid structures. \cite{GAT} adopts attention mechanism in GNNs. \cite{EGGN} proposes an effective way to exploit features of edges in GNNs.  Methods\cite{visual_GNNs_1,visual_GNNs_2,visual_GNNs_3} based on GNNs are also applied to many areas in computer vision, since the effectiveness of these GNNs. In this paper, we adopt a variation of convolution operation in \cite{EGGN} to compute feature of nodes in our graph nueral network.

\noindent\textbf{Temporal Action Proposal Generation.}
The goal of temporal action proposal generation task is to retrieve temporal segments that contain action instance with high recall and precision. Previous methods \cite{wang2014action,oneata2014lear} use temporal sliding window to generate candidate proposals. However, durations of ground truth action instances are various, the duration flexibility are neglected in these methods. Some methods  \cite{SST,sw2,sw3,SSAD} adopt multi-scale anchors to generate proposals, and these methods are similar with the idea in anchor-based object detection. \cite{TAG} proposes Temporal Actionness Grouping (TAG) to output actionness probability for each temporal location over the video sequence using a binary actionness classifier. Then, continuous temporal regions with high actionness score are combined to obtain proposals. This method is effective and simple, but the proposal it generates lacks the confidence for ranking. Recently, \cite{BSN,BMN} generate proposals in a bottom-up and top-down fashion. As bottom-up, boundaries of temporal proposals are predicted at first. As top-down, content between boundaries is evaluated as a confidence score. While the relations between boundaries and content is not utilized explicitly, which is quite important we believe. In this paper, we combine boundary probability predictions and confidence evaluation procedures into a whole by graph neural network. It facilitates information exchanging through these two branches, and brings strong quality improvement in temporal action proposal generation and temporal action detection.

\section{Our Approach}
\begin{figure}
\setlength{\belowcaptionskip}{0cm}
\setlength{\abovecaptionskip}{0.3cm}
\centering
\includegraphics[width=.95 \linewidth]{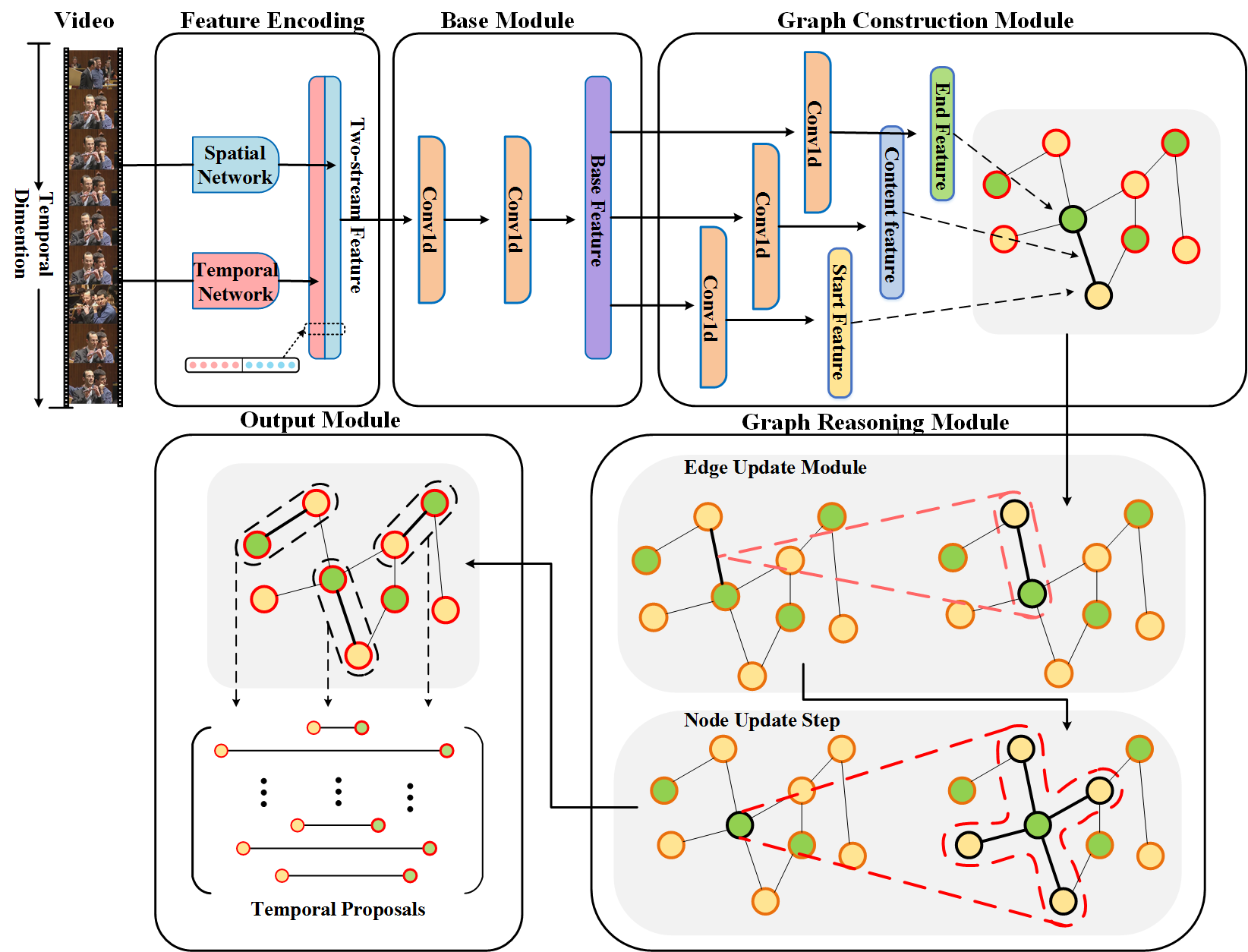}
\caption{The framework of BC-GNN. Feature Encoding encodes the video into sequence of feature. Base Module expands the receptive field. GCM constructs boundary content graph network in which start nodes and end nodes are denoted as green circles and yellow circles separately. GRM updates edges and nodes, to relate information between edges and nodes. Finally, Output Module generates every candidate proposal with each edge and its connected nodes.    }
\label{overview}
\end{figure}

In this section, we will introduce the details of our approach illustrated in Fig.\ref{overview}. In Feature Encoding, visual contents of input video are encoded into feature sequence by a spatial and temporal action recognition network, then this sequence of features is fed into our proposed Boundary Content Graph Neural Network (BC-GNN) framework. There are four modules in BC-GNN: Base Module, Graph Construction Module (GCM), Graph Reasoning Module (GRM) and Output Module. The Base Module is the backbone which is used to exploit local semantic information of input feature sequence. GCM takes feature sequences from Base Module as input and construct a graph neural network. In the GRM module, a new graph computation operation is proposed to update attributes of edges and nodes. Output Module takes the updated edges and nodes as input to predict boundary and content confidence scores. At last, proposals are generated by score fusion and Soft-NMS.

\subsection{Problem Definition}
One untrimmed video consists of a sequence of $l_v$ frames, and this sequence can be denoted as $X={\{x_n\}}^{l_v}_{n=1} $. Action instances in the video content compose a set named $\Psi_g = \{\psi_n = (t_{s}^n,t_{e}^n)\}^{N_g}_{n=1}$, where $t_{s}^n$ and $t_{e}^n$ denote the start and end temporal points of the $n_{th}$  action instance respectively, and $N_g$ denotes the total number of action instances in this video. Classes of these action instances are not considered in temporal action proposal generation task. 

\subsection{Feature Encoding}
Two-stream network \cite{two-stream} is adopted in our framework as visual encoder, since this encoder shows good performance in video action recognition task. This two-stream network consists of spatial and temporal branches. Spatial one is used to encode RGB frames and temporal one is adopted for encoding flow frames. They are designed to capture information from appearance and motion seperately.

More specifically, an input video $X$ with $l_v$ frames is downsampled to a sequence of $l_s$ snippets $S = \{s_n\}_{n=1}^{l_s}$ in a regular temporal interval $\tau$. Thus, the length of snippet sequence $l_s$ is calculated as $l_s = l_v / \tau$. Every snippet $s_n$ in sequence $S$ is composed of a RGB frame $x_n$ and several optical frames $o_n$. After feeding $S$ into two-stream network, two sequences of action class scores are predicted from top layers of both branches. Then, these two sequences of scores are concatenated together at feature dimension to generate a feature sequence $F=\{f_n\}_{n=1}^{l_s}$.

\subsection{Boundary Content Graph Network}

\noindent\textbf{Base Module.}
On one hand, Base Module expands the receptive field, thus it serves as the backbone of whole network. On the other hand, because of the uncertainty of untrimmed videos' length, Base Module applies temporal observation window with fixed length $l_w$ to normalize length of input sequences for the whole framework. The length of observation windows depends on type of datasets. We denote input feature sequence in one window as $F_i \in R^{D_i \times l_w}$, where $D_i$ is the input feature dimension size.

We use two stacked 1D convolution to design our Base Module since local features are needed in sequential parts, written by $F_b = conv1d_{2}(conv1d_{1}(f_i))$. After feeding feature sequence $F_i$ into convolutional layers, $F_b \in R^{D_b \times l_w}$ is generated. 

\begin{figure}
\centering
\setlength{\belowcaptionskip}{-0.cm}
\subfigure[Undirected Graph in GCM]{
\includegraphics[width=.35\linewidth]{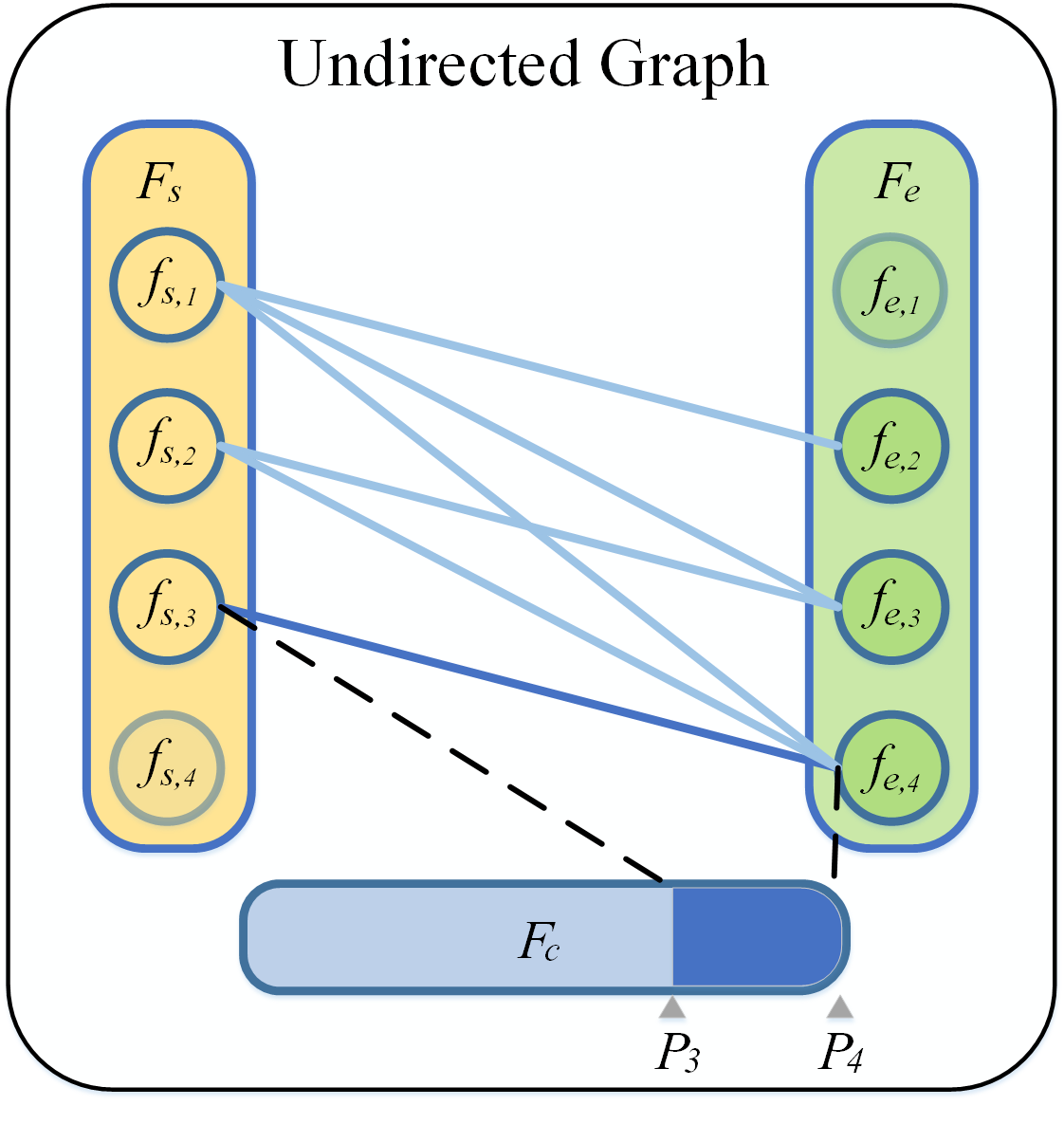}
\label{GCM}
}
\quad\quad\quad\quad
\subfigure[Directed Graph in GRM]{
\label{GRM}
\includegraphics[width=.35\linewidth]{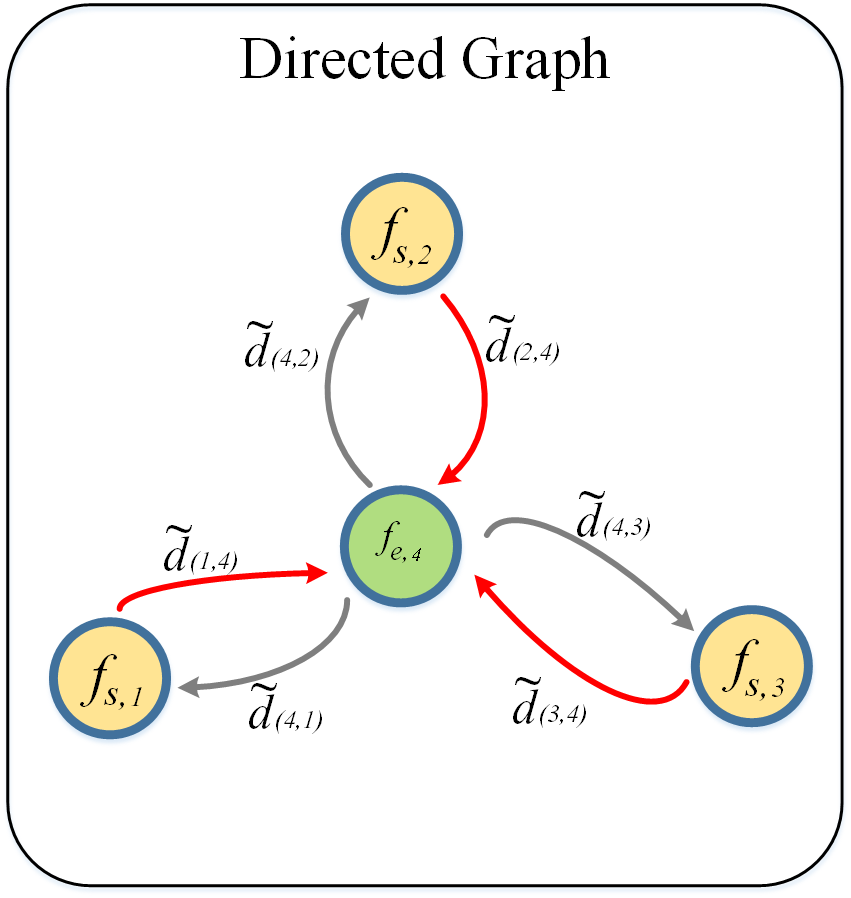}
}

\caption{(a) Construction of undirected graph in GCM. Yellow circle denotes the start node $f_{s,i}$ sampled from feature $F_s$, green circle denotes the end node $f_{e,i}$ sampled from feature $F_e$, and blue line denotes the undirected edge which is generated from feature vectors between temporal locations $P_i$ and $P_j$ in $F_c$. The translucent circles denote the nodes without edge connection.
 (b) Structure of directed graph in GRM. For convenience of description, this digraph only contains one end node and three start nodes. Red curves denote the start to end edge which point from start node to end node, and the grey curves denote the end to start edge which point from end node to start node. }

\end{figure}

\noindent\textbf{Graph Construction Module(GCM).}
The goal of GCM is to construct a boundary content graph network. Fig.\ref{GCM} shows the simplified structure of undirected graph generated by GCM.

Three convolutional layers $conv1d_{s}$, $conv1d_{e}$ and $conv1d_{c}$ will be adopted for $F_b \in R^{D_b \times l_w}$ separately to generate three feature sequence $F_s \in R^{D_g \times l_w}$, $F_e \in R^{D_g \times l_w}$ and $F_c \in R^{D_c \times l_w}$. It should be noted that feature dimension size of $F_s$ and $F_e$ are equal to $D_g$.

 We regard feature sequences $F_s$ and $F_e$ as two sets of feature elements, denoted as $F_s = \{f_{s,i}\}_{i=1}^{l_w}$ and $F_e = \{f_{e,j}\}_{j=1}^{l_w}$, where $f_{s,j}$ and $f_{e,j}$ are the $i_{th}$ start feature in $F_s$ and the $j_{th}$ end feature in $F_e$. Then we conduct the Cartesian product between sets $F_s$ and $F_e$, denoted as $F_s \times F_e = \{(f_{s,i}, f_{s,j})|f_{s,i}\in F_s \land f_{e,j} \in F_e\}$.  To clear out the illegals, we remove every tuple whose start location $i$ is greater than or equal to the end feature location $j$ from the $F_s \times F_e$ and name the start-end pair set to $M_{SE} = \{(f_{s,i}, f_{s,j})|(f_{s,i}\in F_s) \land (f_{e,j} \in F_e) \land (i < j)\}$. The pairs of start and end feature form a start-end pair set $M_{SE}$.

To achieve content representation, we select feature elements between the $i_{th}$ temporal location and the $j_{th}$ location from $F_c$ as a sequence $\{f_{c,n}\}_{n=i}^{j}$. We adopt linear interpolation to achieve constant $N$ vectors at temporal dimension from $\{f_{c,n}\}_{n=i}^{j}$, and denote it as $f_{c,(i,j)} \in R^{D_c \times N}$. After generating $f_{c,(i,j)}$, we reshape its dimension size from ${D_c \times N}$ to $(D_c \cdot N) \times 1$, and apply a fully connected layer $fc_1$ to make dimension size of $f_{c,(i,j)}$ same with $f_{s,i}$ and $f_{e,j}$, denoted as $f_{c,(i,j)} \in R^{D_g}$. Thus, we achieve a content set $M_C = \{f_{c,(i,j)}|i<j\}$. Content between the $i_{th}$ temporal location and the $j_{th}$ temporal location composes content set $M_C$.

Then, the start-end pair set $M_{SE}$ and content set $M_C$ make up a undirected graph. Since the tuple $(f_{s,i}, f_{e,j}) \in M_{SE}$ corresponds to the video segment that starts at the $i_{th}$ temporal location and ends at the $j_{th}$ temporal location. If elements in $F_s$ and $F_e$ are regarded as the nodes of a graph, tuples in $M_{SE}$ identify the connection relationship between these nodes. Meanwhile the tuples in $ M_{SE}$ and elements in $M_C$ are mapped one by one. Therefore, elements in $M_C$ can be regarded as the edges of this graph. Formally, graphs can be denoted as $G=(V,E,I)$, where $V$, $E$ and $I$ are their nodes, edges and incidence functions respectively. In our graph, we define nodes as $V = F_s \cup F_e$, edges as $E = M_C$ and incidence function as $I = M_c \leftrightarrow M_{SE}$, where $M_{SE} \subset V \times V $. We call $f_{s,i}$ start node, and call $f_{e,i}$ end node. 

In summary, we build a restricted undirected bipartite graph in which start nodes are only connected to end nodes whose temporal locations are behind them. It should be noted edge feature in our boundary content graph is not scalars but multi-dimensional feature vectors.

\noindent\textbf{Graph Reasoning Module(GRM).}
In order to enable information exchanging between nodes and edges, we propose a new graph computation operation. One time of graph reasoning operation is applied in a block named Graph Reasoning Block (GRB). GRM consists of two stacked GRBs.

Our graph computation operation is divided into edge update and node update step. Edge update step is intended to aggregate the attributes of the two nodes connected by the edge. As mentioned above, we construct an undirected bipartite graph, in which edges are not directed and start nodes only connect with end nodes. Since the feature required from start nodes to end nodes is different from information from end nodes to start nodes. We converse the undirected graph into a directed graph or a bi-directed edge. This conversion is shown in Fig.\ref{GRM}, every undirected edge is split into two opposite directed edges. In detail, we divide an undirected edge in this graph into two directed edges with the same nodes connection and opposite direction. In other words, one undirected edge turns into two directed edges, which are start to end directed edge and end to start directed edge. We define one directed edge from the $i_{th}$ start feature $f_{s,i} \in F_s$  to the $j_{th}$ end feature $f_{e,j} \in F_e$ as $d_{(i, j)}$, and define directed edge from end feature $f_{e,j}$ to start feature $f_{s,i}$ as $d_{ (j,i)}$, where subscript $i$ is only used for start node, $j$ is only used for end node, and $(i,j)$ identifies the direction of the directed edge which points from the $i_{th}$ start node to the $j_{th}$ end node.

Features of directed edges $d_{(i,j)}$ and $d_{(j,i)}$ are same before the edge updating, denoted as $d_{(i,j)}=d_{(j, i)}=f_{c,(i,j)} $, where $f_{c,(i,j)}$ is feature of the undirected edge in undirected graph. The edge updating can be described as
\begin{equation}
\begin{cases}
 \ \tilde d_ {(i, j)} = \sigma (\theta_{s2e} \times (d_{(i, j)} * f_{s, i} *f_{e, j})) + d_{(i, j)})
 \\
 \ \tilde d_{(j, i)} = \sigma (\theta_{e2s} \times (d_{(j, i)} * f_{s, i} *f_{{e, j}}) + d_{(j, i)})
\end{cases},
\end{equation}
where ``$*$'' and ``$\times$'' denote element-wise product and matrix product separately. $\theta_{s2e} \in R^{D_g \times D_g}$ and $\theta_{e2s} \in R^{D_g \times D_g}$ are different trainable parameter matrices, and ``$\sigma$'' denotes activation function ReLU.

Node update step aims to aggregate attributes of the edges and their adjacent nodes. We adopt the variation of graph convolution in \cite{EGGN}. For the convenience of description, we denote start node and end node as general node $n_k \in R^{D_g}$, where $k$ denotes the $k_{th}$ node in the graph.  The total number of these nodes is $l_N = l_w \cdot 2$, and these general nodes form a set as $N=\{n_k\}_{k=1}^{l_N}$. Meanwhile, we treat updated start to end edge $\tilde d_{(i,j)}$ and updated end to start edge $\tilde d_{(j,i)}$ as general edge $e_{(h, t)} \in R^{D_g} $. These general edges form a set as $E=\{e_{(h,t)}|n_h \in N \land n_t \in N\}$. As usual, the node pointed by the directed edge is called the tail node, and the node where the edge starts is called the head node. It is defined that $e_{(h, t)}$ is from head node $n_h$ to tail node $n_t$. Considering that the number of nodes connected to each other is different, and to avoid increasing the scale of output features through multiplication, we first normalize the features of edges before the graph convolution operation. This normalization operation is described as
\begin{equation}
\tilde e_{(h, t)}^p = \frac{e_{(h, t)}^p}{\sum_{k=1}^{K} e_{(h, k)}^p},
\end{equation}
where $p$ is the $p_{th}$ feature in feature vectors $e_{(h, t)}$ and $ \tilde e_{(h, t)}$, and $K$ is the number of tail nodes. Note that all elements in $e_{(h, t)}$ are nonnegative.
Then the convolution process of node features is described as
\begin{equation}
\tilde n_t = \sigma(\theta_{node} \times (\sum_{h=1}^{H} (\tilde e_{(h, t)} * n_h)) + n_t),
\end{equation}
where trainable matrix $\theta_{node} \in R^{D_g \times D_g}$ is divided into $\theta_{start}$ and $\theta_{end}$ depending on type of node $n_t$, and $H$ is the number of head nodes. This convolution operation gathers the information of head nodes to the tail nodes through the directed edges.

After performing the above two steps, there are a new node feature set $\tilde N = \{\tilde n_k\}_{k=1}^{l_N}$ and an edge feature set $\tilde E = \{\tilde e_{(h,t)}|\tilde n_h \in \tilde N \land \tilde n_t \in \tilde N\}$  generated in one GRB. These two sets become input of the second GRB.

\noindent\textbf{Output Module.}
As shown in Fig.\ref{introduct}, a candidate proposal is generated using a pair of opposite directed edges and their connected nodes. Boundaries and content confidence scores of the candidate proposals are generated based on their nodes and edges, respectively. The details are described as following.

Before fed into Output Module, directed edge feature set $\tilde E$ is divided into a start to end edge feature set and an end to start edge feature set, which are denoted as $\tilde E_{s2e} = \{\tilde e_{s2e,(i, j)}|i<j \land \tilde  e_{s2e} \in \tilde E\}$ and
 $\tilde E_{e2s} = \{
 \tilde e_{e2s,(j, i)},
 |i<j \land \tilde e_{e2s,(j, i)} \in \tilde E\}$.
Meanwhile, node feature set $\tilde N$ is divided into a start node feature set $\tilde N_s = \{\tilde n_{s,i}\}_{i=1}^{l_w}$ and an end node feature set $\tilde N_e=\{\tilde n_{e,j}\}_{j=1}^{l_w}$.
Based on this divided feature sets, we build a candidate proposal feature set
$M_{SCE} =
\{( \tilde n_{s,i}, \tilde n_{e,j},\tilde e_{s2e,(i, j)}, \tilde e_{e2s,(j,i)})
|i<j
\}
$, where $\tilde n_{s,i} \in \tilde N_s$ is the $i_{th}$ start node feature ,
 $\tilde n_{e,j} \in \tilde N_e$ is the $j_{th}$ end node feature,
 $\tilde e_{s2e,(i, j)} \in \tilde E_{s2e}$ is directed edge feature from the $i_{th}$ start node to the $j_{th}$ end node
 and $ \tilde e_{e2s,(j,i)} \in \tilde E_{e2s}$ is directed edge feature from the $j_{th}$ end node to the $i_{th}$ start node. The elements in $M_{SCE}$ are mapped to $M_{SE}$ one by one.

Output Module generates one proposal set $\Psi_p = \{\psi_n\}_{n=1}^{\l_\Psi}$, where $\psi_n = (t_{s}, p_{s}, t_{e}, p_{e}, p_{c})$.  $t_{s}$ and $t_{e}$ are start and end temporal locations of $\psi_n$ separately. $p_{s}$, $p_{e}$ and $p_{c}$ are the confidence scores of boundary locations $t_{s}$, $t_{e}$ and confidence score of content between boundaries $t_{s}$ and $t_{e}$.

Each element in $M_{SCE}$ is computed to get a $\psi_n$, and the computation operation is described as
\begin{equation}
\psi_n =
\begin{cases}
 \ t_{s} = i,
 \\
 \ t_{e} = j,
 \\
 \ p_{s} = \sigma(\theta_{SO} \times \tilde n_{s, i}),
 \\
 \ p_{e} = \sigma(\theta_{EO} \times \tilde n_{e,j}),
 \\
 \ p_{c} = \sigma({\theta_{CO} \times (\tilde e_{s2e, (i, j)} \| \tilde e_{{s2e,(j,i)}})})
\end{cases},
\end{equation}
where ``$\sigma$'' denotes activation function sigmoid, ``$\times$'' denotes matrix multiplication, and ``$\|$'' denotes concatenating operation at feature dimension between vectors. $\theta_{SO}$, $\theta_{EO}$ and $\theta_{CO}$ denote trainable vectors.

\subsection{Training of BC-GNN}

\noindent\textbf{Label Assignment.}
Given a video, we first extract feature sequence by two-stream network \cite{two-stream}. Then, we use sliding observation windows with length $l_w$ in feature sequence to get a series of feature sequences with length of $l_w$.

The ground-truth action instances in this window compose an instance set
$ \Psi_g = \{ \psi_g^n = (t_{g,s}^n, t_{g, e}^n)\}_{n=1}^{l_g} $, where $l_g$ is the size of $\Psi_g$. $\psi_g^n$ starts at the temporal position $t_{g,s}^n$ and ends at $t_{g,e}^n$. For each ground truth action instance $\psi_n^g $, we define its start interval $r_{s}^n = [t_{g,s}^n - d_g^n/10, t_{g, s}^n + d_g^n /10]$ and end interval $r_{g, e}^n = [t_{g,s}^n - d_g^n/10, t_{g, s}^n + d_g^n/10]$ separately, where $d_g^n = t_{g,e}^n - t_{g, s}^n$. After that, the start region and end region are defined as following
\begin{equation}
\begin{cases}
\ r_{g,s} = \mathop{\cup}\limits_{n=1}^{l_g} r_{g,s}^n
\\
\ r_{g,e} = \mathop{\cup}\limits_{n-1}^{l_g} r_{g,e} ^n
\end{cases}
.
\end{equation}

Extracted features in observation window are denoted as $F_i$. Taking $F_i$ as the input, BC-GNN outputs a set $\Psi_p = \{
\psi_n\ =
(t_{s}, p_{s}, t_{e}, p_{e}, p_{c})
\}_{n=1}^{l_p}$, where $l_p$ is the size of $\Psi_p$. Because a plenty of temporal proposals share boundaries, boundary locations $t_s$ and $t_e$ are duplicated in $\Psi_p$. We select a start set $S = \{s_n=(t_s, p_s, b_{s})|\}_{n=1}^{l_s}$, an end set $E = \{e_n=(t_e, p_e, b_e)\}_{n=1}^{l_e}$ and a content set $C =\{c_n = (t_{s}, t_{e}, p_{c}, b_c)\}_{n=1}^{l_c}$ from $\Psi_P$. In these three sets, $b_s$, $b_e$ and $b_c$ are assigned labels for $s_n$, $e_n$ and $c_n$ based on $\Psi_g$. If $t_s$ locates in the scope of $r_{g,s}$, label $b_s$ in start tuple $s_n$ is set to constant 1, otherwise it is set to 0. In the same way we can get the label of $e_n$. If $b_c$ of content tuple $c_n$ is set to 1, two conditions need to be satisfied. One is that $t_s$ and $t_e$ of content tuple $c_n$ located in the regions of $r_{g,s}$ and $r_{g,e}$ respectively. The other is that IoU between $[t_s, t_e]$ and any ground-truth action instances $\psi_g = (t_{g,s}, t_{g, e})$ is larger than 0.5.

\noindent\textbf{Training Objective.}
We train BC-GNN in the form of a multi-task loss function. It can be denoted as
\begin{equation}
L_{objective} = L_{bl}(S) + L_{bl}(E) + L_{bl}(C).
\end{equation}
We adopt weighted binary logistic regression loss function $L_{bl}$ for start, end and content losses, where $L_{bi}$ is denoted as
\begin{equation}
L_{bl}(X) = \sum_{n=1}^{N} (\alpha^{+} \cdot bi \cdot \log p_n + \alpha^{-} \cdot (1-bi)) \cdot \log(1-p_n)),
\end{equation}
where
$\alpha^+ = \frac{N}{\sum(b_i)}$, $\alpha^- =  \frac{N}{\sum(1-b_i)}$ and $N$ is the size of set $X$.

\subsection{Inference of BC-GNN}
During inference, we conduct BC-GNN with same procedures described in training to generation proposals set $\Psi_p = \{\psi_n = (t_s, t_e, p_s, p_e, p_c)\}_{n=l}^{l_p}$. To get final results, BC-GNN undergos score fusion and redundant proposals suppression steps.

\noindent\textbf{Score Fusion.}
To generate a confidence score for each proposal $\psi_n$, we fuse its boundary probabilities and content confidence score by multiplication. This procedure can be described as
\begin{equation}
p_{f} = p_{s} * p_{e} * p_{c}.
\end{equation}
Thus, the proposals set can be denoted as $\Psi_p = \{\psi_n = (t_s, t_e, p_f)\}_{n=l}^{l_p}$.

\noindent\textbf{Redundant Proposals Suppression.}
After generating a confidence score for each proposal, it is necessary to remove redundant proposals which highly overlap with each other. In BC-GNN, we adopt Soft-NMS algorithm to remove redundant proposals. Candidate proposal set $\Psi_P$ turns to be $\Psi'_P = {\psi_n = (ts, te, p'_f)}_{n=1}^{l'_P}$, where $l'_P$ is the number of final proposals.

\section{Experiment}
We present details of experimental settings and evaluation metrics in this section. Then we compare the performance of our proposed method with previous state-of-the-art methods on benchmark datasets.
\subsection{Dataset and Setup}

\noindent\textbf{ActivityNet-1.3.}
This dataset is a large-scale dataset for temporal action proposal generation and temporal action detection tasks. ActivityNet-1.3 contains 19,994 annotated videos with 200 action classes, and it is divided into three sets by ratio of 2:1:1 for training, validation and testing separately.

\noindent\textbf{THUMOS-14.}
This dataset includes 1,010 videos and 1,574 videos in the validation and testing sets with 20 classes. And it contains action recognition, temporal action proposal generation and temporal action detection tasks. For the action proposal generation and detection tasks, there are 200 and 212 videos with temporal annotations in the validation and testing sets.

\noindent\textbf{Evaluation Metrics.}
Average Recall (AR) with Average Number (AN) of proposals per video calculated under different temporal intersection over union (tIoU) is used to evaluate the quality of proposals. AR calculated at different AN is donated as AR@AN. tIoU thresholds [0.5 : 0.05 : 0.95] is used for ActivityNet-1.3 and tIoU thresholds [0.5 : 0.05 : 1.0] is used for THUMOS-14. Specially, the area under the AR vs. AN curve named AUC is also used as an evaluation metric in ActivityNet-1.3 dataset.

Mean Average Precision (mAP) is used to evaluate the results of action detector. Average Precision (AP) of each class is calculated individually. On ActivityNet-1.3 dataset, a set of tIoU thresholds $[0.5: 0.05: 0.95]$ is used for calculating average mAP and tIoU thresholds $\{0.5,0.75,0.95\}$ for mAP. On THUMOS-14, mAP with tIoU thresholds  \{0.3, 0.4, 0.5, 0.6, 0.7\} is used.

\noindent\textbf{Implement Details.}
We adopt two-stream network \cite{two-stream} for feature encoding, which pre-trained on training set of ActivityNet-1.3. The frame interval $\tau$ is set to 5 in THUMOS-14 and 16 in ActivityNet-1.3.
In Base Module, we set the length of observation window  $l_w$ to 128 on THUMOS-14. And in GCM, we get rid of the segments more than 64 snippets, which can cover 98\% of all action instances.
We linearly interpolate feature sequence of each video to $100$ at the temporal dimension in ActivityNet-1.3, which means $lw = 100$ in this dataset. The learning rate of training BC-GNN is set to $0.0001$, and weight decay is set to $0.005$ on both datasets. We conduct $20$ epoch of model training with the strategy of early stopping.

\subsection{Temporal Action Proposal Generation}
Temporal action proposal generation method aims to find segments in videos which highly overlap with ground-truth action instances. We compare BC-GNN with state-of-the-art methods to verify the effectiveness of our method in this section.

\noindent\textbf{Comparison with state-of-the-art methods.}
Comparative experiments are conducted on two widely used benchmarks ActivityNet-1.3 and THUMOS-14.

The results of comparison on validation of ActivityNet-1.3 dataset between our method and other state-of-the-art temporal action proposal generation approaches are shown in Table \ref{anet auc table}. 
Our method BC-GNN outperforms other leading methods by a large margin, and our method performs particularly well in aspect of AR@100.

\begin{table}
\setlength{\abovecaptionskip}{-.5cm}
\setlength{\belowcaptionskip}{-.1cm}
\caption{Comparison between our approach and other state-of-the-art methods on validation set of ActivityNet-1.3 dataset in terms of AR@AN and AUC.}
\setlength{\tabcolsep}{1mm}
\begin{center}
\begin{tabular}{ccccccccc}

\hline
Method       &Prop-SSAD \cite{SSAD} &CTAP \cite{CTAP}  & BSN \cite{BSN}   &MGG \cite{MGG}  &BMN \cite{BMN}  & BC-GNN  \\ \hline
AR@100(val)  & 73.01     & 73.17 & 74.16 & 74.54 & 75.01 & \textbf{76.73} \\
AUC(val)    & 64.40     & 65.72 & 66.17 & 66.43 & 67.10 & \textbf{68.05} \\ \hline

\end{tabular}
\end{center}
\label{anet auc table}
\end{table}

\begin{table}
\setlength{\abovecaptionskip}{-1.7cm}
\caption{Comparison between our approach with other state-of-the-art methods on testing set of THUMOS-14 in terms of AR@AN.}
\setlength{\tabcolsep}{2mm}
\centering
\begin{tabular}{ccccccc}
\hline
Feature & Method        & @50   & @100  & @200      & @500  & @1000 \\ \hline
C3D     & SCNN-prop \cite{SCNN-prop}    & 17.22 & 26.17 & 37.01     & 51.57 & 58.20 \\
C3D     & SST \cite{SST}        & 19.90 & 28.36 & 37.90     & 51.58 & 60.27 \\
C3D     & BSN \cite{BSN} + NMS           & 27.19 & 35.38 & 43.61     & 53.77 & 59.50 \\
C3D     & BSN + Soft-NMS           & 29.58 & 37.38 & 45.55     & 54.67 & 59.48 \\
C3D     & MGG \cite{MGG}         & 29.11 & 36.31 & 44.32     & 54.95 & \textbf{60.98} \\
C3D     & BMN \cite{BMN} + NMS           & 29.04 & 37.72 & 46.79     & 56.07 & 60.96 \\
C3D     & BMN + Soft-NMS           & 32.73 & 40.68 & 47.86     & 56.42 & 60.44 \\ \hline
C3D     & BC-GNN + NMS  & \textbf{33.56} & \textbf{41.20} &  \textbf{48.23}   & 56.54 &
59.76  \\
C3D     & BC-GNN + Soft-NMS  & 33.31 & 40.93 &  48.15   & \textbf{56.62} &
60.41  \\
\hline
2Stream & TAG \cite{TAG}         & 18.55 & 29.00 & 39.61     & -     & -     \\
Flow    & TURN \cite{TURN}         & 21.86 & 31.89 & 43.02     & 57.63 & 64.17 \\
2Stream & CTAP \cite{CTAP}         & 32.49 & 42.61 & 51.97     & -     & -     \\
2Stream & BSN \cite{BSN} + NMS          & 35.41 & 43.55 & 52.23     & 61.35 & 65.10 \\
2Stream & BSN  + Soft-NMS           & 37.46 & 46.06 & 53.21     & 60.64 & 64.52 \\
2Stream & MGG \cite{MGG}          & 39.93 & 47.75 & 54.65     & 61.36 & 64.06 \\
2Stream & BMN \cite{BMN} + NMS          & 37.15 & 46.75 & 54.84     & 62.19 & 65.22 \\
2Stream & BMN + Soft-NMS          & 39.36 & 47.72 & 54.70     & 62.07 & 65.49 \\ \hline
2Stream & BC-GNN + NMS    & \textbf{41.15} & \textbf{50.35} & 56.23     & 61.45 &66.00  \\
2Stream & BC-GNN + Soft-NMS    & 40.50 & 49.60 & \textbf{56.33}     & \textbf{62.80} &\textbf{66.57} \\

\hline
\end{tabular}
\label{thomas auc table}
\end{table}

Comparison between our method and other state-of-the-art proposal generation methods on testing set of THUMOS-14 dataset in terms of AR@AN is demonstrate in Table \ref{thomas auc table}. 
 Flow feature, 2Stream feature and C3D feature are adopt as the input of these methods for ensuring a fair comparison. In this experiment, BC-GNN outperforms other state-of-the-art methods in a large margin.

These experiments verify the effectiveness of our BC-GNN. BC-GNN achieves the significant performance improvement since it makes explicit use of interaction between boundaries and content.

\begin{figure}

\setlength{\belowcaptionskip}{-.5cm}
\centering
\includegraphics[width=.8\linewidth]{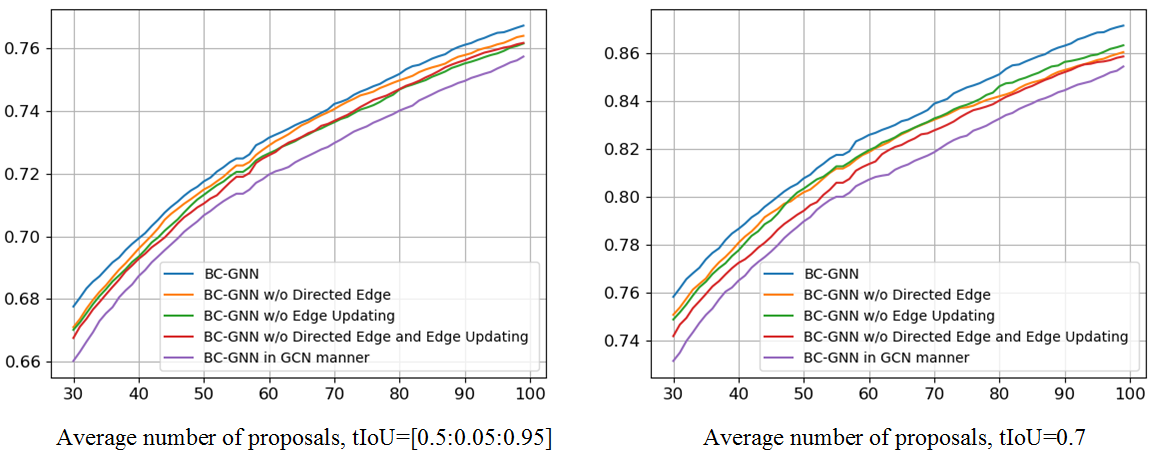}
\caption{Ablation study for our BC-GNN is verified the effectiveness of its modules.}
\label{ablationFig}
\end{figure}

\noindent\textbf{Ablation Study.}
In GRM module, we convert an undirected graph into a directed graph and propose an edge feature updating operation. To evaluate the effectiveness of these strategies, we study ablation experiments in two control groups. We study the models in two control groups. In the first group, we study three types of the graphs: model with Graph Convolutional Network (GCN) manner in which edges are formed by cosine distance between nodes features, and model with directed or undirected edges. Since GCNs does not update edges, the models in the first group do not apply edge updating for the fair. In the second group, we study the effectiveness of directed edge in BC-GNN. The experimental results are listed in Table \ref{ablation} and the average recall against average number of proposals at different tIoU thresholds are shown in Fig.\ref{ablationFig} . The comparison results show that both of strategies are effective and essential.
\begin{table}
\setlength{\abovecaptionskip}{-.5cm}
\setlength{\belowcaptionskip}{-.02cm}
\setlength{\tabcolsep}{2mm}
\caption{Ablation study for model with GCN, edge update step and directed edge.}
\begin{center}
\label{ablation}
\begin{tabular}{ccccc}
\hline
Method  &Directed &Edge updating &AR@100 & AUC(val)\\
\hline
GCN  &-       &- &75.57    & 66.88      \\
BC-GNN &$\times $&$\times$ &76.18  & 67.36     \\
BC-GNN &$\checkmark $&$\times$  &76.15 &67.53      \\\hline
BC-GNN &$\times$&$\checkmark$  &76.40 & 67.79        \\
BC-GNN &$\checkmark$& \checkmark & 76.73 &68.05     \\\hline

\hline
\end{tabular}
\end{center}
\end{table}

\subsection{Temporal Action Detection with Our Proposals}
Temporal action detection is another aspect of evaluating the quality of proposals. On ActivityNet-1.3, we adopt a two-stage framework that detects action instances by classifying proposals. Proposals are generated by our proposal generator firstly and the top-100 temporal proposals per video are retained by ranking. Then, for each video in validation set, its top-1 video-level classification result will be obtained by using two-stream network \cite{anet_cls} and all the proposals of this video share the classification result as their action classes. On THUMOS-14, we use the top-2 video-level classification scores generated by UntrmmedNet \cite{untrimmednets} and proposal-level classification score generated by SCNN-cls to classify first 200 temporal proposals for one video. The results of multiplying the confidence scores of proposals with classification are used for retrieving detection results.

Comparison results between our method and other approaches on validation set of ActivityNet-1.3 in terms of mAP and average mAP are shown in Table \ref{anet_detection}. Our method reaches state-of-the-art on this dataset which validates our approach. We compare our method with other existing approaches on testing set of THUMOS-14 in Table \ref{thumos detection}. Our approach is superior to the other existing two-stage methods on the evaluation metrics mAP, which confirms the effectiveness of our proposed proposal generator.



\begin{table}
\setlength{\abovecaptionskip}{-.4cm}
\setlength{\belowcaptionskip}{.2cm}
\setlength{\tabcolsep}{2mm}
\caption{Action detection results on validation set of ActivityNet-1.3 dataset in terms of mAP and average
mAP.}
\setlength{\tabcolsep}{6mm}
\begin{center}
\label{anet_detection}
\begin{tabular}{ccccc}
\hline
Method       & 0.5    &  0.75   &  0.95   &  Average        \\ \hline
CDC \cite{CDC}         & 43.83  &    25.88  &    0.21   &    22.77          \\
SSN \cite{SNN}         &  39.12  &    23.48  &   5.49   &     23.98          \\
BSN \cite{BSN} + \cite{anet_cls} &   46.45 &  29.96 &  8.02  &  30.03          \\
BMN \cite{BMN} + \cite{anet_cls} &   50.07          &   \textbf{34.78} &   8.29          &   33.85          \\ \hline
BC-GNN +  \cite{anet_cls} &   \textbf{50.56} &   34.75 &   \textbf{9.37}        &   \textbf{34.26} \\ \hline
\end{tabular}
\end{center}
\end{table}

\begin{table}

\setlength{\abovecaptionskip}{-1.5cm}
\setlength{\belowcaptionskip}{.3cm}
\begin{center}

\caption{Comparison between our approach and other temporal action detection methods on THUMOS-14.}
\setlength{\tabcolsep}{4mm}
\label{thumos detection}
\begin{tabular}{ccccccc}

\hline
Method & Classifier & 0.7  & 0.6  & 0.5  & 0.4  & 0.3  \\

\hline
TURN \cite{TURN} & SCNN-cls  & 7.7  & 14.6 & 25.6 & 33.2 & 44.1 \\
BSN \cite{BSN}  & SCNN-cls  & 15.0 & 22.4 & 29.4 & 36.6 & 43.1 \\
MGG \cite{MGG}  & SCNN-cls  & 15.8 & 23.6 & 29.9 & 37.8 & 44.9 \\
BMN \cite{BMN} & SCNN-cls  & 17.0 & 24.5 & 32.2 & 40.2 & 45.7 \\
\hline
BC-GNN & SCNN-cls  & \textbf{19.1}     & \textbf{26.3}     & \textbf{34.2}     & \textbf{ 41.2 }   & \textbf{46.3}     \\
\hline
TURN \cite{TURN} & UNet  & 6.3  & 14.1 & 24.5 & 35.3 & 46.3 \\
BSN \cite{BSN} & UNet  & 20.0 & 28.4 & 36.9 & 45.0 & 53.5 \\
MGG \cite{MGG}  & UNet  & 21.3 & 29.5 & 37.4 & 46.8 & 53.9 \\
BMN \cite{BMN} & UNet  & 20.5 & 29.7 & 38.8 & 47.4 & 56.0 \\
\hline
BC-GNN & UNet  &\textbf{23.1}   & \textbf{31.2}     & \textbf{40.4}     & \textbf{49.1}   & \textbf{57.1}     \\
\hline

\end{tabular}
\end{center}
\end{table}

\section{Conclusion}
In this paper, a new method of temporal action proposal generation named Boundary Content Graph Network (BC-GNN) is proposed. A boundary content graph is proposed to exploit the interaction between boundary probability generation and confidence evaluation. A new graph reasoning operation is also introduced to update the features of nodes and edges in the boundary content graph. In the meantime, an output module is designed to generate proposals using the strengthened features. The experimental results on popular datasets show that our proposed BC-GNN method achieves promising performance in both temporal proposal generation and temporal action detection tasks.
\clearpage

\bibliographystyle{splncs04}
\bibliography{egbib}

\begin{thebibliography}{10}
\providecommand{\url}[1]{\texttt{#1}}
\providecommand{\urlprefix}{URL }
\providecommand{\doi}[1]{https://doi.org/#1}

\bibitem{SST}
Buch, S., Escorcia, V., Shen, C., Ghanem, B., Carlos~Niebles, J.: Sst:
  Single-stream temporal action proposals. In: Proceedings of the IEEE
  Conference on Computer Vision and Pattern Recognition. pp. 2911--2920 (2017)

\bibitem{sw2}
Caba~Heilbron, F., Carlos~Niebles, J., Ghanem, B.: Fast temporal activity
  proposals for efficient detection of human actions in untrimmed videos. In:
  Proceedings of the IEEE Conference on Computer Vision and Pattern
  Recognition. pp. 1914--1923 (2016)

\bibitem{ten}
Carreira, J., Zisserman, A.: Quo vadis, action recognition? a new model and the
  kinetics dataset. In: Proceedings of the IEEE Conference on Computer Vision
  and Pattern Recognition. pp. 6299--6308 (2017)

\bibitem{three}
Dalal, N., Triggs, B., Schmid, C.: Human detection using oriented histograms of
  flow and appearance. In: European Conference on Computer Vision. pp.
  428--441. Springer (2006)

\bibitem{sw3}
Escorcia, V., Heilbron, F.C., Niebles, J.C., Ghanem, B.: Daps: Deep action
  proposals for action understanding. In: European Conference on Computer
  Vision. pp. 768--784. Springer (2016)

\bibitem{six}
Feichtenhofer, C., Pinz, A., Zisserman, A.: Convolutional two-stream network
  fusion for video action recognition. In: Proceedings of the IEEE Conference
  on Computer Vision and Pattern Recognition. pp. 1933--1941 (2016)

\bibitem{CTAP}
Gao, J., Chen, K., Nevatia, R.: Ctap: Complementary temporal action proposal
  generation. In: Proceedings of the European Conference on Computer Vision.
  pp. 68--83 (2018)

\bibitem{TURN}
Gao, J., Yang, Z., Chen, K., Sun, C., Nevatia, R.: Turn tap: Temporal unit
  regression network for temporal action proposals. In: Proceedings of the IEEE
  International Conference on Computer Vision. pp. 3628--3636 (2017)

\bibitem{EGGN}
Gong, L., Cheng, Q.: Exploiting edge features for graph neural networks. In:
  Proceedings of the IEEE Conference on Computer Vision and Pattern
  Recognition. pp. 9211--9219 (2019)

\bibitem{one}
Klaser, A., Marsza{\l}ek, M., Schmid, C.: A spatio-temporal descriptor based on
  3d-gradients (2008)

\bibitem{BMN}
Lin, T., Liu, X., Li, X., Ding, E., Wen, S.: Bmn: Boundary-matching network for
  temporal action proposal generation. In: Proceedings of the IEEE
  International Conference on Computer Vision. pp. 3889--3898 (2019)

\bibitem{SSAD}
Lin, T., Zhao, X., Shou, Z.: Temporal convolution based action proposal:
  Submission to activitynet 2017. arXiv preprint arXiv:1707.06750  (2017)

\bibitem{BSN}
Lin, T., Zhao, X., Su, H., Wang, C., Yang, M.: Bsn: Boundary sensitive network
  for temporal action proposal generation. In: Proceedings of the European
  Conference on Computer Vision. pp. 3--19 (2018)

\bibitem{MGG}
Liu, Y., Ma, L., Zhang, Y., Liu, W., Chang, S.: Multi-granularity generator for
  temporal action proposal. In: Proceedings of the IEEE Conference on Computer
  Vision and Pattern Recognition. pp. 3604--3613 (2019)

\bibitem{oneata2014lear}
Oneata, D., Verbeek, J., Schmid, C.: The lear submission at thumos 2014  (2014)

\bibitem{nine}
Qiu, Z., Yao, T., Mei, T.: Learning spatio-temporal representation with
  pseudo-3d residual networks. In: Proceedings of the IEEE International
  Conference on Computer Vision. pp. 5533--5541 (2017)

\bibitem{Kipf_GCN}
Schlichtkrull, M., Kipf, T.N., Bloem, P., Van Den~Berg, R., Titov, I., Welling,
  M.: Modeling relational data with graph convolutional networks. In: European
  Semantic Web Conference. pp. 593--607. Springer (2018)

\bibitem{two}
Scovanner, P., Ali, S., Shah, M.: A 3-dimensional sift descriptor and its
  application to action recognition. In: Proceedings of the 15th ACM
  International Conference on Multimedia. pp. 357--360. ACM (2007)

\bibitem{visual_GNNs_2}
Shen, Y., Li, H., Yi, S., Chen, D., Wang, X.: Person re-identification with
  deep similarity-guided graph neural network. In: Proceedings of the European
  conference on computer vision. pp. 486--504 (2018)

\bibitem{CDC}
Shou, Z., Chan, J., Zareian, A., Miyazawa, K., Chang, S.: Cdc:
  Convolutional-de-convolutional networks for precise temporal action
  localization in untrimmed videos. In: Proceedings of the IEEE Conference on
  Computer Vision and Pattern Recognition. pp. 5734--5743 (2017)

\bibitem{SCNN-prop}
Shou, Z., Wang, D., Chang, S.: Temporal action localization in untrimmed videos
  via multi-stage cnns. In: Proceedings of the IEEE Conference on Computer
  Vision and Pattern Recognition. pp. 1049--1058 (2016)

\bibitem{two-stream}
Simonyan, K., Zisserman, A.: Two-stream convolutional networks for action
  recognition in videos. In: Advances in neural information processing systems.
  pp. 568--576 (2014)

\bibitem{eight}
Tran, D., Bourdev, L., Fergus, R., Torresani, L., Paluri, M.: Learning
  spatiotemporal features with 3d convolutional networks. In: Proceedings of
  the IEEE International Conference on Computer Vision. pp. 4489--4497 (2015)

\bibitem{GAT}
Veli{\v{c}}kovi{\'c}, P., Cucurull, G., Casanova, A., Romero, A., Lio, P.,
  Bengio, Y.: Graph attention networks. arXiv preprint arXiv:1710.10903  (2017)

\bibitem{four}
Wang, H., Schmid, C.: Action recognition with improved trajectories. In:
  Proceedings of the IEEE International Conference on Computer Vision. pp.
  3551--3558 (2013)

\bibitem{wang2014action}
Wang, L., Qiao, Y., Tang, X.: Action recognition and detection by combining
  motion and appearance features. THUMOS14 Action Recognition Challenge
  \textbf{1}(2), ~2 (2014)

\bibitem{five}
Wang, L., Xiong, Y., Wang, Z., Qiao, Y.: Towards good practices for very deep
  two-stream convnets. arXiv preprint arXiv:1507.02159  (2015)

\bibitem{untrimmednets}
Wang, L., Xiong, Y., Lin, D., Van~Gool, L.: Untrimmednets for weakly supervised
  action recognition and detection. In: Proceedings of the IEEE conference on
  Computer Vision and Pattern Recognition. pp. 4325--4334 (2017)

\bibitem{visual_GNNs_3}
Wang, X., Gupta, A.: Videos as space-time region graphs. In: Proceedings of the
  European conference on computer vision. pp. 399--417 (2018)

\bibitem{SNN}
Xiong, Y., Zhao, Y., Wang, L., Lin, D., Tang, X.: A pursuit of temporal
  accuracy in general activity detection. arXiv preprint arXiv:1703.02716
  (2017)

\bibitem{visual_GNNs_1}
Yan, S., Xiong, Y., Lin, D.: Spatial temporal graph convolutional networks for
  skeleton-based action recognition. In: Thirty-second AAAI conference on
  artificial intelligence (2018)

\bibitem{TAG}
Zhao, Y., Xiong, Y., Wang, L., Wu, Z., Tang, X., Lin, D.: Temporal action
  detection with structured segment networks. In: Proceedings of the IEEE
  International Conference on Computer Vision. pp. 2914--2923 (2017)

\bibitem{anet_cls}
Zhao, Y., Zhang, B., Wu, Z., Yang, S., Zhou, L., Yan, S., Wang, L., Xiong, Y.,
  Lin, D., Qiao, Y., et~al.: Cuhk \& ethz \& siat submission to activitynet
  challenge 2017. arXiv preprint arXiv:1710.08011  (2017)

\end{thebibliography}
\end{document}